\documentclass[letterpaper]{article} 
\usepackage{aaai25}  
\usepackage{times}  
\usepackage{helvet}  
\usepackage{courier}  
\usepackage[hyphens]{url}  
\usepackage{graphicx} 
\usepackage{natbib}  
\usepackage{caption} 
\usepackage{algorithm}
\usepackage{algorithmic}
\usepackage{xcolor}
\usepackage{newfloat}
\usepackage{listings}
\DeclareCaptionStyle{ruled}{labelfont=normalfont,labelsep=colon,strut=off} 
\floatstyle{ruled}
\newfloat{listing}{tb}{lst}{}
\floatname{listing}{Listing}
\nocopyright 

\usepackage{booktabs}
\usepackage{amsmath}
\usepackage{multirow}

\title{VISION: Robust and Interpretable Code Vulnerability Detection \\ Leveraging Counterfactual Augmentation}

\author{
    David Egea\textsuperscript{\rm 1}\textsuperscript{\rm 2}, 
    Barproda Halder\textsuperscript{\rm 1}, 
    Sanghamitra Dutta\textsuperscript{\rm 1}
}
\affiliations {
    \textsuperscript{\rm 1}University of Maryland College Park\\
    \textsuperscript{\rm 2}Universidad Pontificia Comillas\\
    davidegea@alu.comillas.edu, bhalder@umd.edu, sanghamd@umd.edu
}

\begin{document}

\maketitle

\begin{abstract}
Automated detection of vulnerabilities in source code is an essential cybersecurity challenge, underpinning trust in digital systems and services. Graph Neural Networks (GNNs) have emerged as a promising approach as they can learn the structural and logical code relationships in a data-driven manner. However, the performance of GNNs is severely limited by training data imbalances and label noise. GNNs can often learn ``spurious'' correlations due to superficial code similarities in the training data, leading to detectors that do not generalize well to unseen real-world data. In this work, we propose a new unified framework for robust and interpretable vulnerability detection---that we call VISION---to mitigate spurious correlations by systematically augmenting a counterfactual training dataset. Counterfactuals are samples with minimal semantic modifications that have opposite prediction labels. Our complete framework includes: (i) generating effective counterfactuals by prompting a Large Language Model (LLM); (ii) targeted GNN model training on synthetically paired code examples with opposite labels; and (iii) graph-based interpretability to identify the truly crucial code statements relevant for vulnerability predictions while ignoring the spurious ones. We find that our framework reduces spurious learning and enables more robust and generalizable vulnerability detection, as demonstrated by improvements in overall accuracy (from 51.8\% to 97.8\%), pairwise contrast accuracy (from 4.5\% to 95.8\%), and worst-group accuracy increasing (from 0.7\% to 85.5\%) on the widely popular Common Weakness Enumeration (CWE)-20 vulnerability. We also demonstrate improvements using our proposed metrics, namely, intra-class attribution variance, inter-class attribution distance, and node score dependency. We provide a new benchmark for vulnerability detection, CWE-20-CFA, comprising 27,556 samples from functions affected by the high-impact and frequently occurring CWE-20 vulnerability, including both real and counterfactual examples. Furthermore, our approach enhances societal objectives of transparent and trustworthy AI-based cybersecurity systems through interactive visualization for human-in-the-loop analysis.
\end{abstract}

\renewcommand\thefootnote{}
\footnotetext{This is the authors’ preprint version. The paper has been accepted for publication at AIES 2025. Copyright is held by the Association for the Advancement of Artificial Intelligence (AAAI).}
\renewcommand\thefootnote{\arabic{footnote}}

\section{Introduction}

Software vulnerabilities remain a primary entry point for cyberattacks, making early and accurate detection essential for securing modern digital systems~\cite{chernis2018mlmethods}. The deep syntactic and semantic structure of code remains difficult for traditional static and dynamic analysis tools to model, restricting their applicability and effectiveness~\cite{yamaguchi2014cpg}. To this end, Graph Neural Networks (GNNs)~\cite{scarselli2008thegnnmodel, wang2023finegrainedvulnerabilitygnn} offer a promising approach for vulnerability detection by representing source code as graphs that capture syntax, control flow, and data dependencies. Architectures such as Devign~\cite{zhou2019devign} exemplify the potential of GNN-based methods to automatically learn complex program semantics and support data-driven vulnerability detection.

However, the effectiveness of GNN-based detectors is constrained by the quality of the datasets they are trained on. Popular benchmarks often suffer from label duplication, class imbalance, and inconsistent or noisy labeling~\cite{ding2024vulnerabilitydetectioncodelanguage, guo2023qualityissuesvulndatasets, croft2023dataqualitysoftwarevulnerability}, which can lead to spurious correlations. Spurious correlations are formally defined as statistical associations between input features and target labels that do not reflect a true causal relationship; instead, they arise from coincidental patterns or confounding variables in the data~\cite{haig2003spurious,ye2024spuriouscorrelationsmachinelearning, bell2024multipledimensionsspuriousnessmachine, steinmann2024navigatingshortcutsspuriouscorrelations,gala,halder2024quantifying}. Models influenced by such correlations may appear accurate during training but often fail to generalize, as they rely on misleading or dataset-specific signals. These limitations are further compounded by the fact that GNN-based detectors behave as opaque boxes, offering limited interpretability and transparency into their decision-making process. Simply predicting vulnerabilities is not enough, as it remains unclear \emph{what aspects of the input does the model rely on and if those aspects are really crucial}. Without this understanding, it becomes difficult to ensure robust and trustworthy decisions.

To address these limitations, we propose \textbf{VISION} (\textbf{V}ulnerability \textbf{I}dentification and \textbf{S}puriousness mitigation via counterfactual augmentat\textbf{ION}), a unified framework for robust and interpretable vulnerability detection by systematically augmenting a counterfactual dataset. In our context, we define \textit{counterfactuals} as minimally altered code examples whose vulnerability labels are inverted (e.g., from vulnerable to benign or vice versa), inspired from counterfactual explanations  in tabular classification~\cite{wachter2018counterfactualexplanationsopeningblack}. As a simple illustration, consider the transformation of a line like \texttt{strcpy(dest, "fixed\_string");} into \texttt{strcpy(dest, user\_input);}. While syntactically similar, the latter introduces a potential vulnerability by incorporating unvalidated input. \emph{By generating such minimally modified examples during training, our framework encourages GNN models to better distinguish genuine vulnerability patterns from spurious ones.} To support this, we train a GNN model inspired by the Devign architecture~\cite{zhou2019devign} on these counterfactual pairs, and then integrate the Illuminati explainer~\cite{he2023illuminatiexplaininggraphneural} to identify \emph{the most influential code statements} via subgraph-based attributions. 

We evaluate our framework on a challenging large-scale vulnerability detection dataset called the Common Weakness Enumeration (CWE)-20~\cite{mitre_cwe20}, consisting of  Improper Input Validation---a high-impact vulnerability ranked 4th in the 2022 CWE Top 25---highlighting our potential to enhance robustness and generalization in critical security contexts. We find that counterfactual augmentation leads to substantial improvements across key robustness and generalization metrics---including pairwise accuracy (rising from 4.5\% to 95.8\%) and worst-group accuracy (from 0.7\% to over 85\%)---demonstrating our effectiveness in mitigating spurious correlations and enhancing model generalization.

Our framework also includes an interactive visualization module for attributing the most influential statements in the source code (explainable inspection) for human-in-the-loop settings. Explanations generated under our framework tend to prioritize semantically relevant code regions, suggesting reduced reliance on spurious patterns. In essence, our framework VISION is a significant advancement in  explainability for vulnerability detection, that can potentially help practitioners understand why code is flagged as risky, enabling more reliable remediation~\cite{sharma2022xaiforcybersecurity}.

To summarize, our key contributions are: 
\begin{itemize}
    \item \textbf{Novel counterfactual data augmentation strategy} leveraging Large Language Models (LLMs) to improve the robustness of GNN-based vulnerability detection by effectively mitigating spurious correlations.
    \item \textbf{Empirical validation} on a challenging vulnerability detection dataset called CWE-20, demonstrating substantial improvements on several generalization metrics, showcasing a scalable way to improve performance without relying on new data sources.
    \item \textbf{Extensive benchmark} that we call CWE-20-CFA constructed by generating counterfactual examples from existing CWE-20 samples.
    \item  \textbf{Visualization module} for qualitative analysis of the model's decision-making, showing more semantically-meaningful input source code attributions. 
\end{itemize}

\section{Related Work}

\subsubsection{Vulnerability Detection with Machine Learning.} 
Early machine learning methods used handcrafted features like token frequencies and structural cues but struggled to generalize across codebases \cite{neuhaus2007vulture, chernis2018mlmethods}. Deep learning improved detection leveraging RNNs for sequential dependencies \cite{ziems2021securityvulnerabilitydetectionusing}, CNNs for local patterns \cite{wu2022vulcnn}, and transformers like CodeBERT and GraphCodeBERT for long-range context \cite{feng2020codebertpretrainedmodelprogramming, guo2021graphcodebert}. GNNs model source code as structured graphs \cite{scarselli2008thegnnmodel, zhou2019devign, yamaguchi2014cpg}, thus enhancing robustness. LLMs have also been directly applied for cross-language and general-purpose vulnerability detection \cite{sultana2024codevulnerabilitydetectioncomparative, zhou2024largelanguagemodelvulnerability}.

\subsubsection{Dataset Quality and Challenges.} 
Label noise is a critical problem, with mislabeling rates reported between 20–71\% in several benchmark datasets \cite{croft2023dataqualitysoftwarevulnerability}. Duplication rates from 17–99\% can cause overfitting to surface-level patterns, while class imbalance biases models toward non-vulnerable classes \cite{guo2023qualityissuesvulndatasets}. Sample similarity---where inter-class examples are too alike, or intra-class examples are too diverse---further complicates learning \cite{liu2022vulnerabilitydatasets}. Recent datasets like PrimeVul \cite{ding2024vulnerabilitydetectioncodelanguage} introduce stricter de-duplication and validation. Others, such as DiverseVul \cite{chen2023diversevul}, aim to increase linguistic and structural diversity, improving generalization. Yet high-quality, well-balanced vulnerability datasets remain scarce, posing a significant barrier to developing robust detection models.

\subsubsection{Data Augmentation Strategies for Source Code.} 
Augmentation methods for vulnerability detection aim to preserve both syntactic validity and semantic meaning. Techniques include CodeGraphSMOTE~\cite{ganz2023codegraphsmote}, which interpolates in graph latent space to balance datasets, and transformation-based methods that diversify code while preserving labels~\cite{liu2024enhancingcodevulnerabilitydetection}. LLM-based approaches like VulScribeR~\cite{daneshvar2024exploringragbasedvulnerabilityaugmentation} offer scalable synthesis of vulnerable code. Outside of the source code domain, other related works~\cite{kaushik2020learningdifferencemakesdifference, ross2021explainingnlpmodelsminimal, dissanayake2024model,temraz2021solvingclassimbalanceproblem,ding2022data,kong2022robust,liu2022graph} have explored strategies for augmenting synthetic examples in the dataset, e.g., finding counterfactuals via optimization for numeric data, manually creating new examples for text data, oversampling underrepresented groups, or perturbation-based techniques. However, counterfactual augmentation for code vulnerability detection---focusing on minimal, label-flipping edits leveraging an LLM---remains largely unexplored.


\subsubsection{Explainability in Vulnerability Detection.}
Model-agnostic tools like SHAP and LIME explain feature contributions \cite{lundberg2017shap, ribeiro2016lime}. In code analysis, IVDetect \cite{li2021ivdetect} leverages program dependency graphs to localize vulnerabilities. CFExplainer \cite{chu2024cfexplainer} generates counterfactuals to highlight decision boundaries. GNNExplainer and PGM-Explainer provide subgraph-based explanations for GNNs \cite{ying2019gnnexplainer, vu2020pgmexplainerprobabilisticgraphicalmodel}. Illuminati~\cite{he2023illuminatiexplaininggraphneural} is a domain-specific explainer for GNNs in cybersecurity that identifies the most influential nodes, edges, and attributes in a prediction, producing interpretable subgraphs that clarify the model’s decision-making process. Notably, Illuminati only reveals the underlying decision logic \emph{given a specific model, and does not mitigate spuriousness or improve model performance.} Instead, our work unifies spuriousness mitigation and explanation in a single framework VISION to identify the truly crucial parts of the source code. While we incorporate Illuminati as an internal explainer, we complement it with an interactive, human-in-the-loop visualization module, enabling users to inspect attributions on the same input in a user-friendly manner, essential for real-world applicability.

\section{Proposed Framework: VISION}

\begin{figure}
  \centering
  \includegraphics[width=\linewidth]{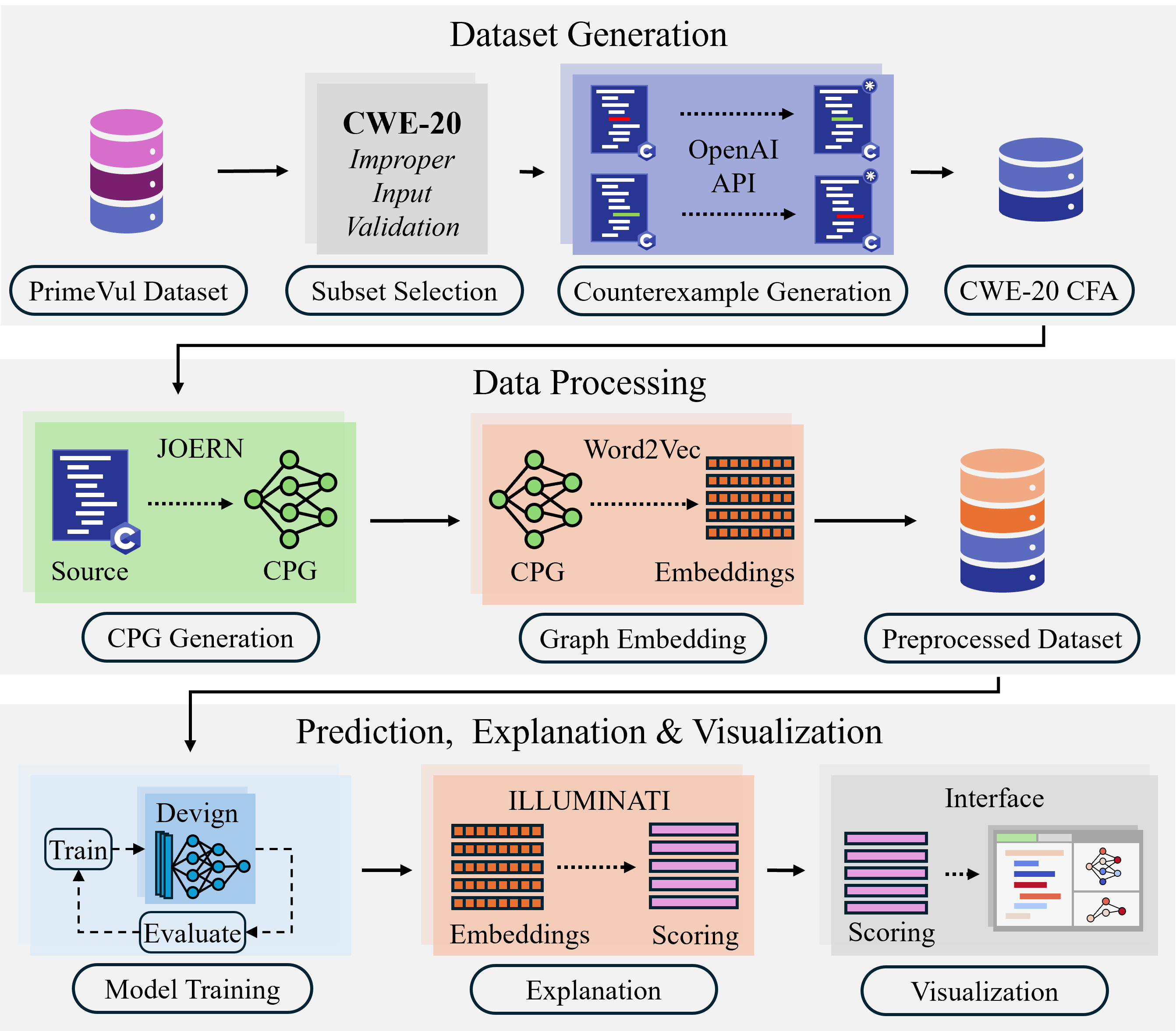}
  \caption{Complete architecture of our VISION Framework. Starting from the original PrimeVul dataset, the framework illustrates the end-to-end process: dataset filtering to CWE-20, counterfactual generation and class balancing, graph construction via Joern-generated CPGs, embedding extraction, model training using Devign, and finally, explanation generation with Illuminati and integration of a visualization module for interpretability.}
  \label{fig:vision_framework_architecture}
\end{figure}

This section presents the VISION framework~\footnote{The implementation of the VISION framework is publicly available
at \url{https://github.com/David-Egea/VISION}.}, designed to enhance the robustness and interpretability of vulnerability detection systems. Focusing on the CWE-20 vulnerability from the PrimeVul~\cite{ding2024vulnerabilitydetectioncodelanguage}, our framework generates counterfactuals---minimally edited examples with opposite labels---by leveraging LLMs to balance the training data and improve generalization. Code samples are parsed into Code Property Graphs (CPGs) using Joern, and then embedded for input to the Devign model. Interpretability is achieved through the Illuminati explainer, with an additional interactive module supporting qualitative analysis of the model’s input attributions. Figure~\ref{fig:vision_framework_architecture} illustrates the full architecture.

\subsection{Dataset Selection: CWE-20 Vulnerability}

To evaluate our framework VISION under realistic conditions, we derive our training and evaluation data from PrimeVul~\cite{ding2024vulnerabilitydetectioncodelanguage}, a recently released high-quality vulnerability dataset for code language models. This dataset was curated with human-verified labels and rigorous de-duplication, resulting in lower label noise and high diversity of real-world code patterns. 

We focus exclusively on examples of \textbf{CWE‑20 \textit{(Improper Input Validation)}} (see~\cite{mitre_cwe20}). This vulnerability class, characterized by the failure to properly sanitize or validate external inputs, was chosen for three main reasons:

\begin{enumerate}
    \item \textbf{Clarity of semantics}. Improper input validation vulnerabilities often exhibit clear and recognizable coding patterns, such as missing boundary checks or unchecked buffer lengths, which facilitate both automated augmentation and human analysis.
    \item \textbf{Data availability}. PrimeVul contains a sufficient number of CWE‑20 instances ($\sim$14k) to support both model training and reliable statistical evaluation without resorting to over‑sampling or synthetic over‑generation.
    \item \textbf{Real-world relevance}. CWE‑20 vulnerabilities remain among the most commonly exploited in practice, reinforcing the need for effective detection to strengthen software security~\cite{cwe2022top25}.
\end{enumerate}

\begin{figure}[tb]
  \centering
  \includegraphics[width=\linewidth]{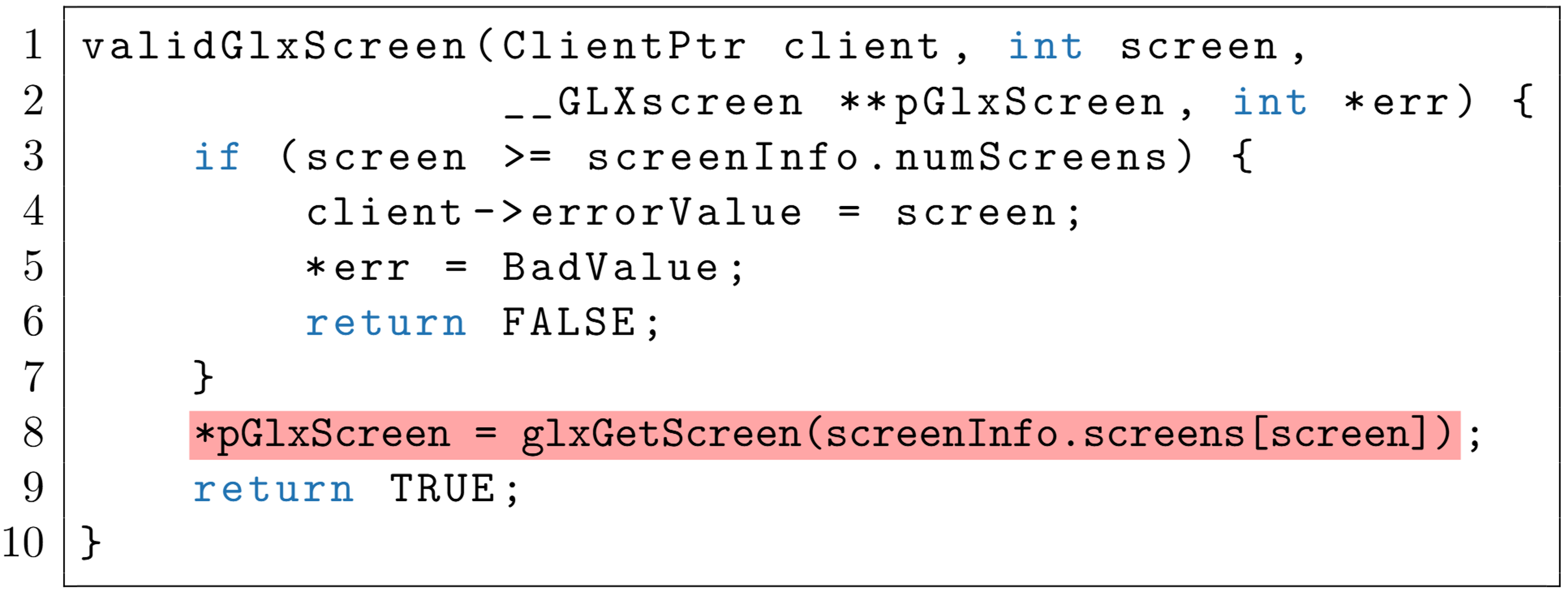}
  \caption{
  CWE-20 Improper Input Validation example. The function \texttt{validGlxScreen} checks whether the input \texttt{screen} is greater than or equal to the number of available screens, but fails to validate negative values. This omission allows invalid array access and illustrates a classic case of insufficient input validation.
  }
  \label{fig:cwe20_example_vul}
\end{figure}

A simple, illustrative example from the CWE‑20 dataset is shown in Figure~\ref{fig:cwe20_example_vul}. The function \texttt{validGlxScreen} takes a \texttt{screen} index and returns a pointer to its data. However, it only checks whether \texttt{screen} is too large, neglecting negative values (or other invalid ranges), and thus allowing invalid dereferences.

\subsubsection{Generalization to Other Vulnerabilities and Languages.}
Although this study focuses on C source code examples of CWE-20 for illustrative purposes, VISION is a general-purpose, dataset-agnostic framework. Applying it to other CWEs (e.g. CWE‑416 Use‑After‑Free or CWE‑787 Out‑of‑Bounds Write), or even to a mixture of vulnerabilities, would essentially follow the same process: (1) assemble a corpus of code examples labeled with the target vulnerability where any initial class imbalance is acceptable; (2) leverage an LLM to generate counterfactuals that equalize the dataset; (3) translate all snippets into CPGs using an appropriate parser (Joern currently provides front‑ends for twelve widely used languages); and (4) discard any invalid generated graphs. The downstream graph‑learning and attribution components can operate without modification.

\subsection{Counterfactual Generation and Augmentation}

\subsubsection{Definition of Counterfactuals.}

First, we define a \textit{counterfactual} as a minimally modified version of a source code function whose vulnerability label is flipped relative to the original. These modifications preserve syntactic and semantic validity while altering the vulnerability, transforming a benign sample into a vulnerable one, or vice versa. This concept draws inspiration from counterfactual explanations for tabular classification~\cite{wachter2018counterfactualexplanationsopeningblack,verma2020counterfactual,hamman2023robustcounterfactual,dutta2022robust} where counterfactual explanations are the closest point on the other side of the decision boundary. It also has intuitive connections with contrastive and counterfactual learning, where the goal is to expose the model to near-boundary examples~\cite{kaushik2020learningdifferencemakesdifference, ross2021explainingnlpmodelsminimal, dissanayake2024model,temraz2021solvingclassimbalanceproblem}.

Formally, given a fixed predictive model $f(\cdot)$, input graph $G = (V, E)$ with node set $V = \{v_1, v_2, \ldots, v_N\} $ and edge set $E = \{ (v_i, v_j) \mid v_i, v_j \in V \} $ , original prediction $f(G) = y$ and counterfactual prediction $y_c \ne y$, a counterfactual mapping $ e : G \rightarrow G'$ such that $f(e(G)) = y_c$.

\begin{figure*}[tb]
  \centering
  \includegraphics[width=0.75\linewidth]{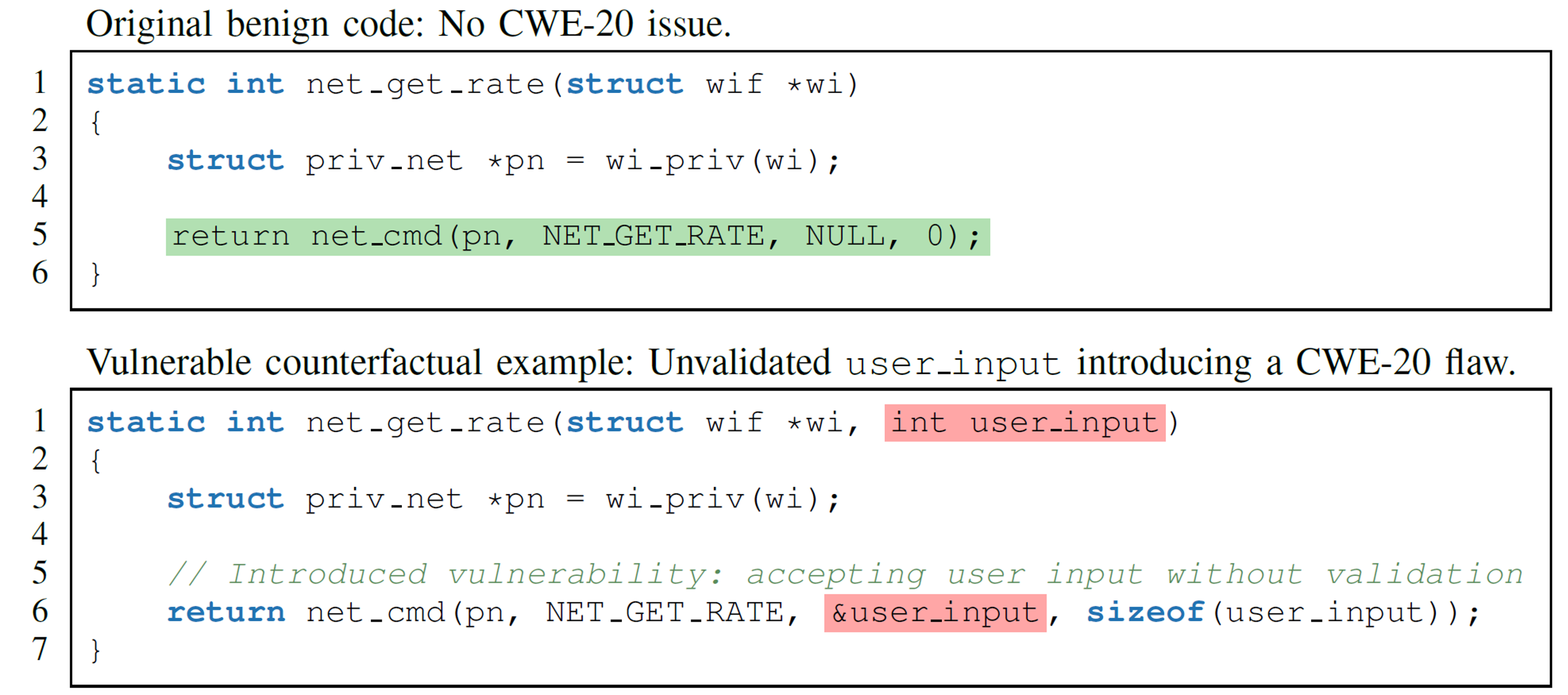}
  \caption{Illustration of a counterfactual code pair used in data augmentation. The top function is benign, safely invoking \texttt{net\_cmd} with no external input. The bottom version introduces a CWE-20 (Improper Input Validation) vulnerability by replacing a fixed argument with user-provided input (\texttt{user\_input}) that is passed without validation.
  }
  \label{fig:counterfactual_code_examples}
\end{figure*}

We observe that counterfactuals are particularly useful in the context of vulnerability detection, because: (i) they balance the dataset by providing an equal number of examples with opposite label; and (ii) they help the model learn to distinguish subtle input validation changes that mark the presence (or absence) of a vulnerability (see~Figure~\ref{fig:counterfactual_code_examples}), leading to an improved understanding of the truly crucial aspects of the input. Unlike traditional augmentation methods that inject noise or generate synthetic samples from scratch, counterfactuals maintain high fidelity to real-world code, improving both training stability and downstream robustness. 

\subsubsection{Counterfactual Generation Strategy.}

We generate the counterfactuals using a prompt-based rewriting approach with the GPT-4o-mini model via the OpenAI API. Each source function was minimally modified to flip its vulnerability label, either introducing or removing a CWE-20 vulnerability, based on a prompt dynamically built from the function’s original source code, label and CWE type.

Figure~\ref{fig:counterfactual_code_examples} shows an example of a counterfactual transformation. The original benign function (top) calls \texttt{net\_cmd} with a fixed, safe parameter (\texttt{NULL}), ensuring no user input is involved. In the counterfactual version (bottom), a new parameter \texttt{user\_input} is introduced and directly passed to \texttt{net\_cmd} without validation. This edit changes the function's vulnerability status, injecting a CWE-20 Improper Input Validation flaw. Such pairs expose the model to near-identical source code that differ only in vulnerability semantics, encouraging it to learn discriminative patterns.

\subsubsection{Dataset Balancing and Processing Methodology.}

To prevent the model from biasing towards the original or majority-class samples, we explicitly balance the dataset during the augmentation process. First, we extracted only \textbf{CWE‑20 (Improper Input Validation)} examples from the full PrimeVul dataset, filtering it from 224,533 total functions down to a focused subset of 14,944 CWE‑20 samples. This subset included 14,473 benign and only 471 vulnerable examples, revealing a substantial class imbalance. 

To address this imbalance, we applied counterfactual generation to both benign and vulnerable samples. This strategy ensured that we were not over-representing either original or synthetic instances, effectively doubling the dataset while producing a perfectly balanced distribution of classes. Problematic or incorrectly generated examples as well as original samples whose counterfactuals could not be reliably created or validated were removed, resulting in the final dataset: CWE-20-CFA, consisting of 27,556 samples (13,778 original functions and 13,778 counterfactuals), equally divided between the two classes (Table~\ref{tab:dataset_balance}). 

\begin{table}[tb]
\centering
\begingroup
\fontsize{9pt}{12pt}\selectfont
\begin{tabular}{lccc}
\toprule
\textbf{Dataset Stage} & \textbf{Benign} & \textbf{Vulnerable} & \textbf{Total} \\
\midrule
PrimeVul                    & 218,529 & 6,004  & 224,533 \\
CWE-20 PrimeVul             & 14,473  & 471    & 14,944  \\
\textbf{CWE-20 CFA} & \textbf{13,778}  & \textbf{13,778} & \textbf{27,556}  \\
-- \textit{Original}             & 13,349 & 429    & 13,778 \\
-- \textit{Counterfactual}       & 429    & 13,349 & 13,778 \\
\bottomrule
\end{tabular}
\endgroup
\caption{CWE-20 Dataset Filtering and Balancing Summary}
\label{tab:dataset_balance}
\end{table}

Code samples were converted into graph representations to enable training within a GNN framework. This transformation was performed using Joern, a state-of-the-art static analysis tool that constructs Code Property Graphs (CPGs) by unifying multiple structural views of the code \cite{yamaguchi2014cpg}. Each function was parsed into a CPG and serialized into a structured graph format. A Word2Vec-style embedding encoder was then applied to map code tokens and edge types into continuous feature vectors. 

\subsection{Base Model for Vulnerability Detection}

The VISION framework builds upon the \textbf{Devign} architecture, a Graph Neural Network (GNN)-based architecture specifically designed for vulnerability detection in software source code \cite{zhou2019devign}. Devign operates on Code Property Graphs (CPGs) \cite{yamaguchi2014cpg}, placing emphasis on the Abstract Syntax Tree (AST) to model the syntactic structure of source code. While ASTs can identify simple issues like insecure arguments, their combination with control flow graphs (CFG) and data flow graphs (DFG) enables the model to cover a wider range of vulnerabilities and learn patterns effectively.

Devign architecture includes three primary components:

\begin{enumerate}
    \item \textbf{Graph Embedding Layer.} Converts source code into a composite graph structure, capturing multiple semantic dimensions.
    \item \textbf{Gated Graph Recurrent Layers (GGRU).} Extract meaningful features by iteratively propagating and aggregating information through graph nodes.
    \item \textbf{Convolutional (Conv) Module.} Performs graph-level classification by effectively summarizing node embeddings into predictive vulnerability labels.
\end{enumerate}

The original Devign model was extensively evaluated on manually labeled datasets from large-scale, open-source C projects such as the Linux Kernel, QEMU, Wireshark, and FFmpeg. The evaluation demonstrated Devign’s significant improvement over state-of-the-art methods, achieving an average accuracy improvement of 10.51\% and F1-score improvement of 8.68\%~\cite{zhou2019devign}. 

VISION uses Devign as the baseline model structure due to its proven ability to encode intricate semantic structures and effectively classify vulnerabilities at a granular level.

\subsection{Visualization Module}

We developed a visualization module to support interpretability and human-in-the-loop analysis, with a particular focus on inspecting individual source code examples and explaining the behavior of the trained model through input attributions. Its primary goal is to provide evidence for the central hypothesis of this work: that counterfactual-based augmentation leads to improved model accuracy, robustness, and a better understanding of source code through more semantically meaningful attributions.

This module is built on top of the Illuminati explainer \cite{he2023illuminatiexplaininggraphneural} and displays model predictions, confidence scores, and node-level importance values. Illuminati leverages Devign  as the underlying representation for generating explanations, revealing the model's decision-making process through minimal and sufficient subgraph extraction. Predictions are color-coded---green when the model predicts correctly, and red when it misclassifies---providing immediate visual feedback on performance. 

One of the key features of our visualization module is the highlighted source code view, where the original function text is color-coded based on the importance scores derived from the model explanation. A continuous color scale maps higher importance to warmer colors, allowing users to quickly identify which statements the model considered most influential. In parallel, the graph view represents the function as a map of nodes and edges that follow the program's logical flow, applying the same color scale for interpretability consistency. 

Beyond direct attribution visualization, the system also offers subgraph-based interpretability. This includes positive subgraphs, where nodes are incrementally added in order of importance until the model recovers the correct prediction; negative subgraphs, where nodes are progressively removed to identify the minimal structural weaknesses that lead to prediction failure; and optimal subgraphs, which represent the subset of nodes that yield the highest overall confidence in the prediction.

\begin{figure*}[tb]
\centering
\includegraphics[width=0.83\linewidth]{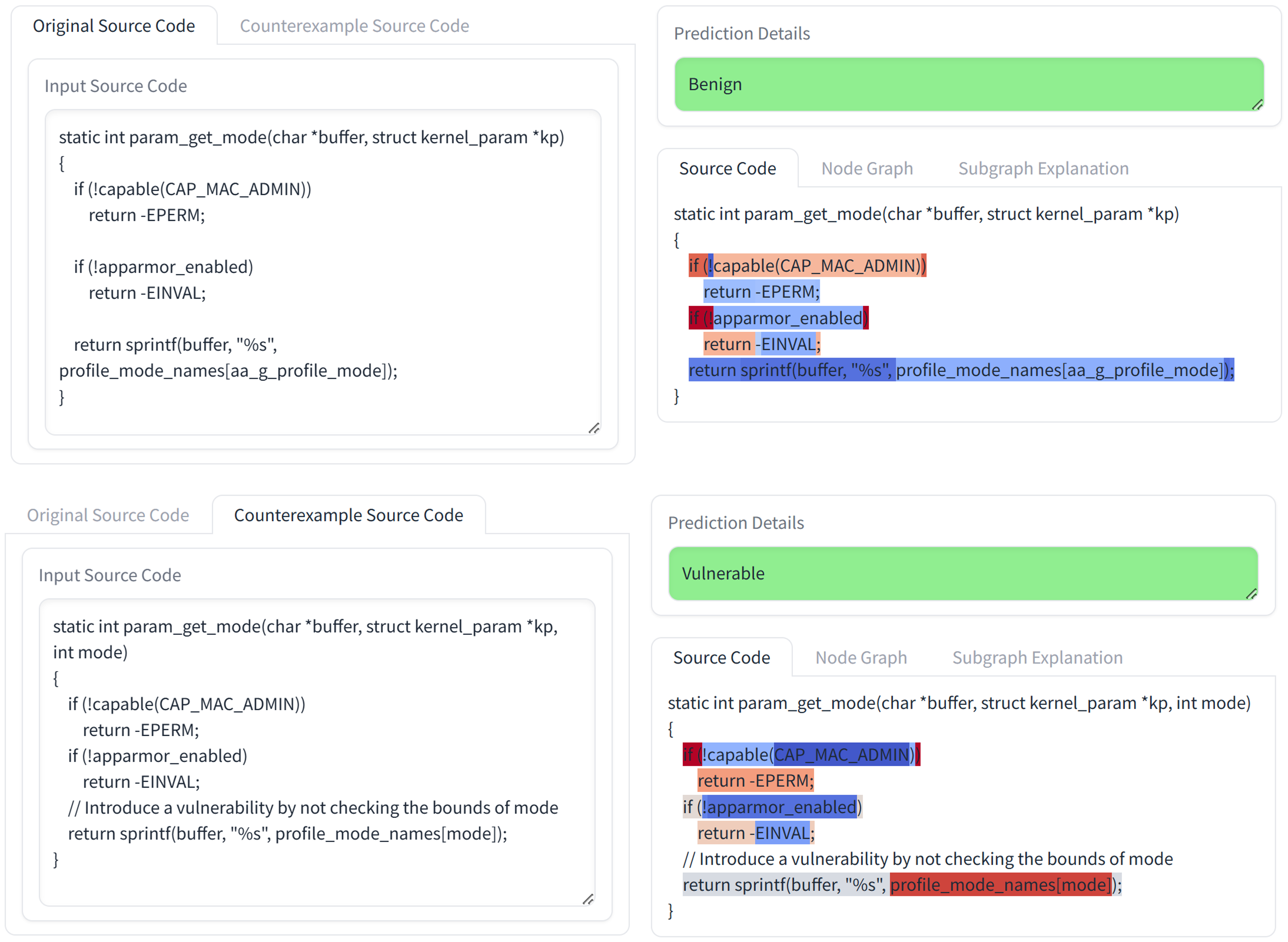}
\caption{Integrated visualization module showcasing model predictions and explanation scores for an original benign function (top) and its vulnerable counterfactual (bottom). The right-hand side highlights tokens with attribution scores, where red intensity indicates higher importance. The module enables intuitive inspection of the model's input attribution shifts, demonstrating how the counterfactual structure influences both the prediction and explanation.
}
\label{fig:visualize_orig}
\end{figure*}

An example of this process is shown in Figure~\ref{fig:visualize_orig}, which compares the visualization output for an original benign function (top) and its corresponding vulnerable counterfactual (bottom). The highlighted tokens and graph regions clearly demonstrate how the model’s attention shifts in response to the augmented structure.

While the visualization module is not the core contribution, it is a powerful tool for qualitative inspection, debugging, and explanation verification. It has proven particularly useful in validating whether the model’s input attributions align with semantically relevant regions of the code, especially for conducting ablation studies with and without using our proposed counterfactual-based augmentation.

\section{Experimental Evaluation}

To evaluate the effectiveness of our counterfactual augmentation framework, we analyze model performance across a series of training benchmarks with varying proportions of original and counterfactual source code examples. \emph{Evaluation is conducted using both conventional metrics, such as accuracy, precision, recall, and F1-score, as well as a set of metrics specifically aimed at evaluating robustness and spuriousness mitigation. These metrics include pair-wise accuracy, worst-group accuracy, causal effect detection, embedding space neighborhood analysis, intra-class attribution variance, inter-class attribution distance, and a new node score dependency metric introduced in this work.}

\subsection{Experimental Setup}

\subsubsection{Dataset Splitting and Augmentation Strategy.}

To evaluate the impact of counterfactual-based augmentation, we construct benchmark datasets with varying ratios of original to counterfactual functions. All benchmarks are derived from the CWE-20-CFA dataset generated using the VISION framework. Each dataset contains the same total number of examples, balanced between benign and vulnerable classes, but differs in the original-to-counterfactual composition.

A fixed, balanced test set is used across all benchmarks to ensure consistent evaluation. The full dataset is first partitioned by unique IDs using an 80/10/10 train-validation-test split. Each ID is assigned exclusively to either the training or test set to preserve the integrity of original–counterfactual pairs. The test set contains both versions of each function, ensuring perfect class balance and mirrored pairings.

\begin{figure}[tb]
  \centering
  \includegraphics[width=\linewidth]{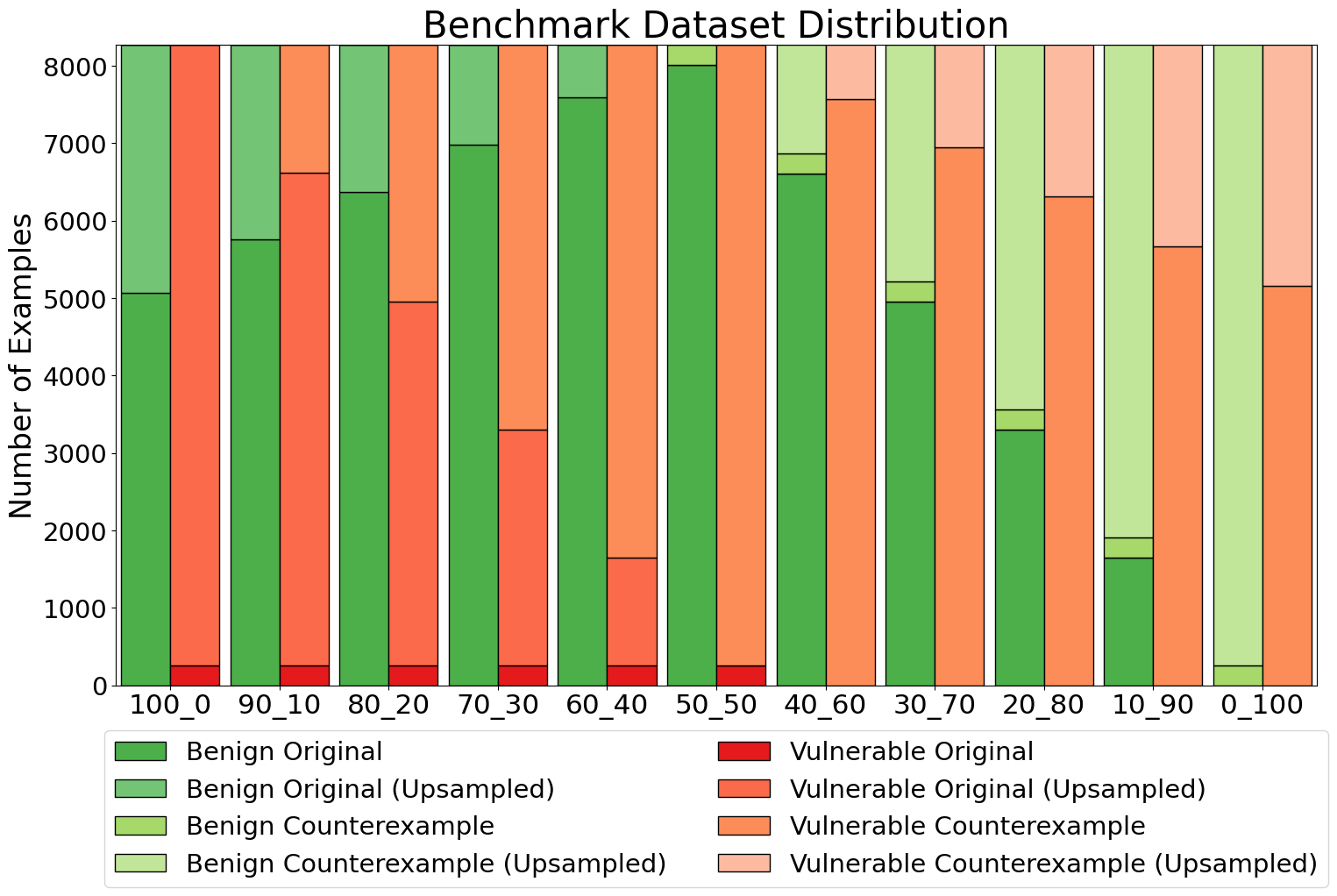}
  \caption{Benchmark dataset distribution by class and data source (original vs. counterfactual). Each bar stack shows the number of examples for benign and vulnerable samples, and the split between original, counterfactual, and upsampled sources.}
  \label{fig:benchmark_distribution_plot}
\end{figure}

Training benchmarks range from 100\% original to 100\% counterfactual, in 10\% increments. To maintain label balance within each split, examples are independently upsampled per class and data source as needed. This setup allows the model to learn from subtle syntactic and semantic differences, improving its ability to distinguish between benign and vulnerable code. Figure~\ref{fig:benchmark_distribution_plot} summarizes the dataset composition. The test dataset is kept constant for all evaluations to ensure fair comparison and consists of an equal number of benign and vulnerable functions, with each original sample paired with its corresponding counterfactual.

\subsection{Performance Across Benchmarks}

We report standard evaluation metrics—accuracy, precision, recall, and F1-score---for models trained under varying original/counterfactual data ratios, ensuring fair comparison through a consistent test set. As shown in Table~\ref{tab:benchmark_results_table}, incorporating counterfactual examples consistently improves model performance across all evaluation metrics. The 100/0 configuration yields perfect precision but suffers from extremely low recall and F1-score, highlighting severe overfitting to superficial patterns. In contrast, performance peaks around the 50/50 split, where the model demonstrates strong generalization with balanced precision and recall. However, relying exclusively on synthetic data, as in the 0/100 case, leads to notable performance degradation---indicating that synthetic examples alone lack sufficient information for effective learning. Overall, these results confirm that balanced integration of counterfactuals enhances learning robustness while maintaining predictive stability.

\begin{table}[tb]
\centering
\setlength{\tabcolsep}{6pt}
\begingroup
\fontsize{9pt}{11pt}\selectfont
\begin{tabular}{lcccc}
\toprule
\textbf{Split} & \textbf{Accuracy} & \textbf{Precision} & \textbf{Recall} & \textbf{F1-score} \\
\midrule
100/0   & 0.518 & 1.000 & 0.036 & 0.069 \\
90/10   & 0.867 & 0.996 & 0.737 & 0.847 \\
80/20   & 0.955 & 0.960 & 0.948 & 0.954 \\
70/30   & 0.970 & 0.960 & 0.980 & 0.970 \\
60/40   & \textbf{0.978} & 0.961 & \textbf{0.997} & \textbf{0.979} \\
50/50   & 0.960 & 0.957 & 0.962 & 0.960 \\
40/60   & 0.970 & \textbf{0.998} & 0.941 & 0.969 \\
30/70   & 0.951 & 0.949 & 0.953 & 0.951 \\
20/80   & 0.930 & 0.904 & 0.962 & 0.932 \\
10/90   & 0.919 & 0.875 & 0.978 & 0.924 \\
0/100   & 0.799 & 0.726 & 0.957 & 0.826 \\
\bottomrule
\end{tabular}
\endgroup
\caption{Performance across original/counterfactual training splits on the balanced test set.}
\label{tab:benchmark_results_table}
\end{table}

\subsection{Robustness and Spurious Correlation Analysis}


In the context of source code, spurious correlations often emerge when a model learns to associate vulnerability labels with superficial patterns that do not reflect the true semantics or security posture of the code. These misleading correlations often stem from recurring harmless syntax patterns, rather than true indicators of security flaws.

\begin{figure}[tb]
\centering
\includegraphics[width=\linewidth]{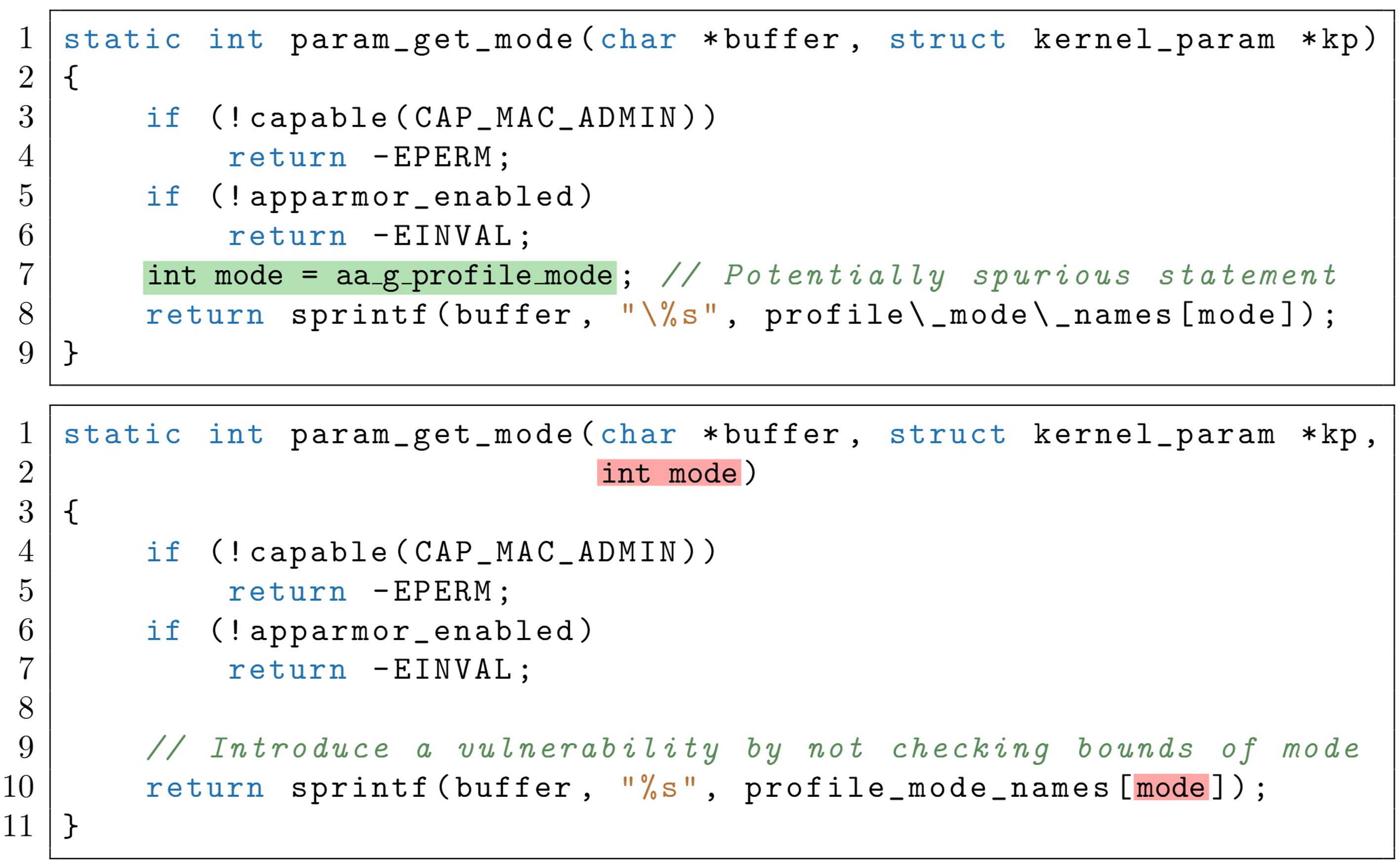}
\caption{
Illustration of spurious correlation in source code. The upper (benign) function assigns a safe internal value to \texttt{mode}, while the lower (vulnerable) version takes \texttt{mode} as unchecked user input. Without sufficient counterfactuals, a model may incorrectly associate the presence of the \texttt{mode} variable with safe behavior, failing to recognize its misuse in the vulnerable case.
}
\label{fig:spurious_corr_code}
\end{figure}

Figure~\ref{fig:spurious_corr_code} illustrates this phenomenon with two semantically related functions. The upper function is benign and retrieves a configuration value (\texttt{aa\_g\_profile\_mode}) from a secure internal source. The lower function is vulnerable, as it accepts a user-controlled input (\texttt{mode}) that is directly passed to \texttt{sprintf()} without bounds checking. A model trained only on examples like the benign variant may erroneously learn that the presence of the mode variable is indicative of safe behavior---due to its association with sanitized sources---thereby failing to flag the vulnerable pattern when mode originates from external input.

By training on both the original and counterfactual versions of such examples, the model is exposed to minimal but security-critical changes in the code structure. This encourages learning that is grounded in true semantic distinctions, rather than dataset-specific shortcuts, and is key to mitigating spurious correlations in source code analysis.

\subsubsection{Pair-Wise Accuracy.}

\begin{table*}[tb]
\centering
\setlength{\tabcolsep}{4pt}
\fontsize{9pt}{11pt}\selectfont
\begin{tabular}{lcccccccccccccc}
\toprule
\textbf{Split} & \textbf{P-C} & \textbf{P-V} & \textbf{P-B} & \textbf{P-R} & \textbf{WGA2} & \textbf{WGA3} & \textbf{WGA4} & \textbf{WGA5} & \textbf{WGA6} & \textbf{WGA7} & \textbf{Purity} & \textbf{Intra-B} & \textbf{Intra-V} & \textbf{Inter-D} \\
\midrule
100/0   & 4.50  & 0.00  & 95.43 & 0.07 & 0.0171 & 0.0096 & 0.0073 & 0.0067 & 0.0058 & 0.0067 & 0.707 & 0.01103 & 0.01027 & 0.00061 \\
90/10   & 74.09 & 1.38  & 23.88 & 0.65 & 0.7309 & 0.7156 & 0.7052 & 0.7126 & 0.5909 & 0.5952 & 0.907 & 0.01120 & 0.01035 & 0.00073 \\
80/20   & 91.07 & 5.44  & 3.27  & 0.22 & \textbf{0.9115} & \textbf{0.8828} & \textbf{0.8757} & 0.8512 & 0.8444 & 0.8205 & 0.953 & 0.01096 & 0.01046 & 0.00027 \\
70/30   & 94.63 & 4.86  & 0.36  & 0.15 & 0.9056 & 0.8745 & 0.8757 & \textbf{0.8595} & 0.8444 & 0.8205 & 0.962 & 0.01109 & 0.00995 & 0.00010 \\
60/40   & 93.69 & 6.31  & 0.00  & \textbf{0.00} & 0.9056 & 0.8745 & 0.8703 & 0.8512 & \textbf{0.8444} & 0.8205 & \textbf{0.967} & 0.01134 & 0.01030 & 0.00010 \\
\textbf{50/50} & \textbf{95.79} & \textbf{0.44} & \textbf{0.00} & 3.77 & 0.8991 & 0.8667 & 0.8555 & 0.8087 & 0.7955 & 0.8095 & 0.944 & \textbf{0.01061} & \textbf{0.01030} & \textbf{0.00160} \\
40/60   & 94.12 & 1.02  & 4.50  & 0.36 & 0.8471 & 0.8089 & 0.8092 & 0.7739 & 0.7955 & 0.8067 & 0.966 & 0.01122 & 0.01036 & 0.00017 \\
30/70   & 87.52 & 8.13  & 3.85  & 0.51 & 0.8777 & 0.8400 & 0.8266 & 0.7739 & 0.7727 & 0.7857 & 0.941 & 0.01101 & 0.01010 & 0.00038 \\
20/80   & 70.97 & 27.72 & 1.02  & 0.29 & 0.8820 & 0.8622 & 0.8497 & 0.8174 & 0.8182 & \textbf{0.8333} & 0.929 & 0.01144 & 0.01036 & 0.00028 \\
10/90   & 77.94 & 20.54 & 0.65  & 0.87 & 0.8584 & 0.8350 & 0.8152 & 0.8265 & 0.8149 & 0.8099 & 0.910 & 0.01103 & 0.01046 & 0.00008 \\
0/100   & 41.51 & 57.40 & 0.65  & 0.44 & 0.5398 & 0.4983 & 0.5030 & 0.4966 & 0.4754 & 0.4742 & 0.856 & 0.01122 & 0.01007 & 0.00099 \\
\bottomrule
\end{tabular}
\caption{Comprehensive evaluation across training splits, covering robustness, generalization, and explanation quality. Metrics include: pair-wise agreement—P-C (correct contrast), P-V (both predicted vulnerable), P-B (both predicted benign), and P-R (flipped predictions); higher P-C and lower P-V/P-B/P-R indicate better discrimination. Worst-Group Accuracy (WGA, $k{=}2$–$7$): higher is better, reflects subgroup robustness. Neighborhood Purity: higher values indicate stronger class consistency in the embedding space and better semantic separation. Attribution metrics include Intra-class Attribution Variance (lower is better, measures consistency) and Inter-class Attribution Distance (higher is better, reflects class separability).}
\label{tab:spurious_metrics_table}
\end{table*}

Pair-wise accuracy, introduced in \cite{ding2024vulnerabilitydetectioncodelanguage}, measures how well a model distinguishes between pairs of semantically similar functions with opposite vulnerability labels, typically an original and its minimally modified counterfactual.

Formally, a high pair-wise classification accuracy indicates that the model distinguishes between subtle code-level changes that cause label flips, rather than relying on spurious features. This metric is an indicator of the model's sensitivity to subtle, meaningful changes in source code.

Pair-wise accuracy includes four components: P-C (Pair-Correct) measures correct contrast between original and counterfactual examples; P-V (Pair-Vulnerable) and P-B (Pair-Benign) represent incorrect predictions where both functions are classified as vulnerable or benign, respectively; and P-R (Pair-Reversed) captures flipped predictions. High P-C and low P-V, P-B, and P-R indicate the model effectively distinguishes subtle, vulnerability-inducing changes without relying on spurious patterns.

As shown in Table~\ref{tab:spurious_metrics_table}, the 50/50 benchmark achieves the highest correct contrast (P-C = 95.79\%), indicating that a balanced mix of original and counterfactual data helps the model focus on subtle but important vulnerability indicators. Extremes like 100/0 or 0/100 result in higher confusion and misclassification, suggesting that spurious correlations dominate when training is biased toward only one data type.

\subsubsection{Worst-Group Accuracy.}

Worst-Group Accuracy (WGA) is a robustness metric that captures a model’s weakest performance across latent subgroups defined by both structural code patterns and class labels \cite{idrissi2022simpledatabalancingachieves}. It reflects how well the model avoids overfitting to spurious correlations and maintains reliable predictions even for hard-to-classify regions of the data.

Since no explicit spurious attributes are available, we adopt an unsupervised approach to define these subgroups. First, we extract latent code embeddings from the trained model and apply K-means clustering to identify groups of structurally or stylistically similar functions. Each function is then assigned to a subgroup based on its cluster ID and ground truth label. Groups with fewer than 1\% of the total data are discarded to avoid instability. The WGA is computed as the lowest classification accuracy among the remaining subgroups.

Given a dataset $\mathcal{D} = \{(x_i, y_i)\}_{i=1}^N$ and a trained model $f$, we define latent groups via unsupervised clustering. Let $g_i \in \mathcal{G}$ denote the group assignment of sample $x_i$, determined by the intersection of its K-means cluster ID and true label $y_i$.

For each group $g \in \mathcal{G}$ with size $|g| > 0.01N$, we compute its accuracy:
\[
\text{Acc}(g) = \frac{1}{|g|} \sum_{i \in g} [f(x_i) = y_i]
\]

The Worst-Group Accuracy (WGA) is then defined as:
\[
\text{WGA} = \min_{g \in \mathcal{G},\; |g| > 0.01N} \text{Acc}(g)
\]

Table~\ref{tab:spurious_metrics_table} presents the WGA values across different original/counterfactual training splits and clustering granularities. The 100/0 model performs poorly across all k-values, with WGA below 2\%, indicating brittle generalization and overfitting to spurious patterns. Models trained with moderate counterfactual integration (e.g., 80/20 to 60/40) achieve the highest and most stable WGA values, consistently above 85\%. Performance slightly drops for fully synthetic or highly imbalanced configurations (0/100 or 50/50). Because \(k=2\) almost perfectly separates the two ground-truth classes, \(\text{WGA}_2\) is invariably the largest and tracks the class-level Purity metric closely (Pearson \(r \approx 0.88\) across splits). For \(k>2\) each class is subdivided into rarer stylistic clusters, so the worst-group accuracy naturally falls. Splits with the highest \(\text{WGA}_2\) also show the lowest intra-class attribution variance. 
\emph{These results support the hypothesis that counterfactual augmentation mitigates spurious correlations by improving the model's consistency across diverse code structures.}

\subsubsection{Neighborhood Analysis in Embedding Space.}

\begin{figure*}[tb]
\centering
\includegraphics[width=\linewidth]{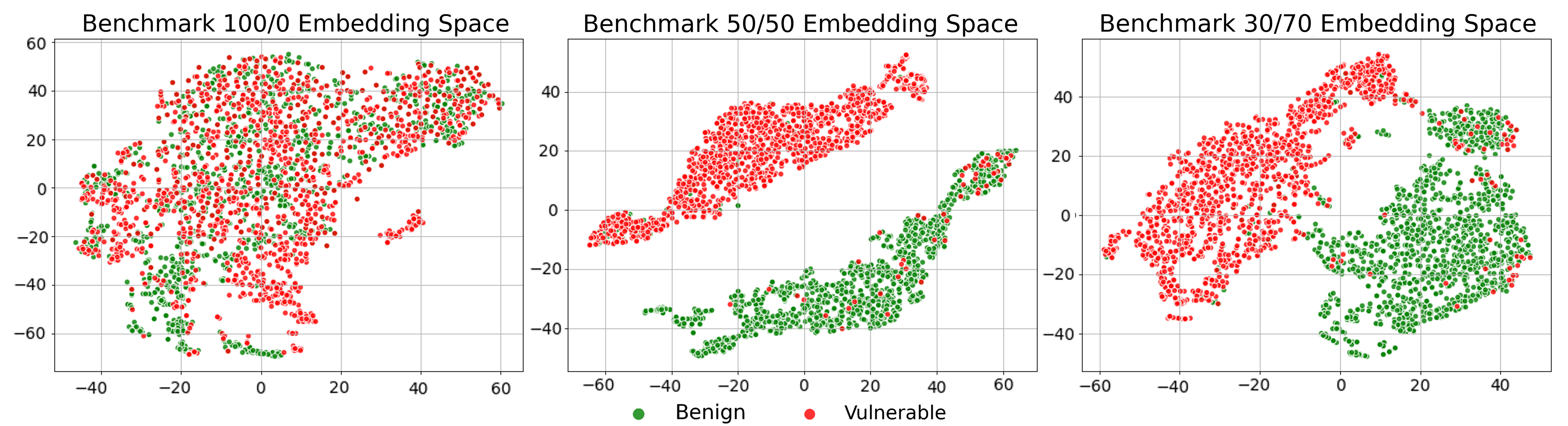}
\caption{t-SNE projections of graph embeddings for benchmarks 100/0 (left), 50/50 (center), and 30/70 (right). Green points represent benign samples, and red points represent vulnerable samples. These visualizations illustrate how different original/counterfactual splits influence the spatial distribution and separation of vulnerability classes.}
\label{fig:tsne_embeddings}
\end{figure*}

Neighborhood purity measures the extent to which embeddings of samples cluster with others of the same class in latent space. Higher purity indicates stronger semantic consistency and suggests that the model has learned to organize representations based on meaningful vulnerability patterns rather than spurious shortcuts. 

We compute the Neighborhood Purity Score using a k-nearest neighbor (kNN) strategy over graph-level embeddings, and report the average class consistency across all samples.

The calculated scores for all training splits appear in Table~\ref{tab:spurious_metrics_table} to show how class consistency in the embedding space changes with different levels of counterfactual augmentation

Neighborhood purity increases sharply as counterfactuals are introduced, peaking around the 60/40 split. The 100/0 model shows the lowest purity (0.71), suggesting under-structured embeddings due to overfitting. While purity remains high for most augmented settings, it slightly declines for heavily synthetic configurations (e.g., 0/100), indicating reduced semantic clustering. These results suggest that moderate augmentation improves embedding structure without inducing shortcut learning.

To further analyze how embedding space evolves under different augmentation ratios, we visualize graph-level representations using t-SNE projections for selected benchmarks (Figure~\ref{fig:tsne_embeddings}). The 0/100 model shows heavily entangled clusters, with benign and vulnerable examples largely mixed---indicating poor separation and potential reliance on non-semantic features. In contrast, the 50/50 benchmark yields a much cleaner separation between classes, suggesting that balanced counterfactual training enables the model to form more meaningful latent representations. The 30/70 exhibits intermediate behavior: vulnerable and benign examples are not as distinctly partitioned as in the 50/50 configuration. This visualization qualitatively supports the neighborhood purity findings and highlights how counterfactual augmentation enhances the model’s ability to semantically organize code representations in latent space.

\subsubsection{Intra-Class Attribution Variance and Inter-Class Attribution Distance.}

In this work, we also propose two other attribution-based metrics to evaluate the consistency and discriminability of model explanations.

The metrics are as follows:

\begin{itemize}
    \item \textbf{Intra-Class Attribution Variance} measures how stable the explainer attributions are across different samples of the same class. High variance suggests that the model may be relying on sample-specific or potentially spurious patterns, while low variance indicates more consistent reasoning.
    \item \textbf{Inter-Class Attribution Distance} quantifies the difference between the average attribution vectors of benign and vulnerable samples. A larger distance suggests better separability between the explanation patterns of both classes.
\end{itemize}

These metrics are computed using attribution vectors generated by the Illuminati explainer. They reflect not only how a model performs, but also how it reasons---providing a more nuanced evaluation of explanation quality, model behavior, and robustness to dataset bias.

As shown in Table~\ref{tab:spurious_metrics_table}, intra-class variance remains relatively stable across benchmarks, but the 50/50 configuration achieves the highest inter-class attribution distance---indicating the clearest semantic separation in model reasoning. This supports that balanced augmentation not only boosts predictive accuracy but also improves explanation quality by helping the model focus on truly semantically relevant and discriminative features.

\subsubsection{Node Score Dependency.}

\begin{figure*}[tb]
\centering
\includegraphics[width=\linewidth]{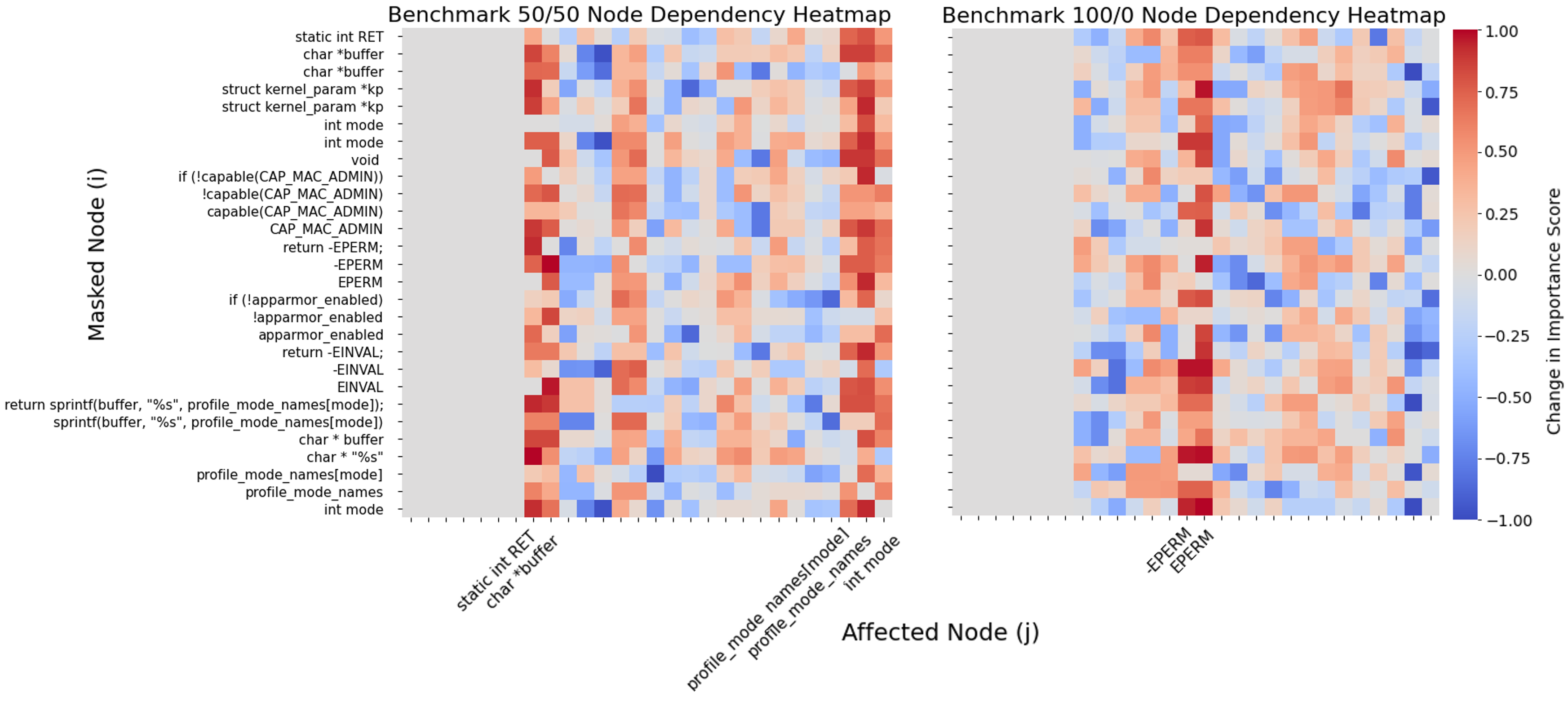}
\caption{Node Score Dependency heatmaps for the same vulnerable function under two training regimes: Left: Benchmark 100/0 and Right: Benchmark 50/50. Each heatmap shows how masking one node (rows) affects the attribution scores of others (columns). Red indicates increased influence; blue indicates reduced importance. The 100/0 model exhibits high focus on spurious context nodes, while 50/50 shows more meaningful attribution alignment.}
\label{fig:node_dependency_comparison}
\end{figure*}

Lastly, we introduce Node Score Dependency, a novel metric to analyze how the importance of each node in a graph depends on others in the context of GNN-based explanation. This metric captures inter-node attribution dynamics by quantifying how attribution scores change when a specific node is removed.

For each node $i$, we temporarily mask or remove it from the graph and recompute the attribution scores using the post hoc explainer. The difference in the importance of any other node $j \ne i$ reflects the degree to which node $j$'s contribution relies on node $i$. This yields a node dependency matrix $M \in \mathrm{R}^{n \times n}$, where:

\[
M_{i,j} = \left| \text{score}_j^{\text{orig}} - \text{score}_j^{(i\text{-removed})} \right|
\]

Here, $M_{i,j}$ represents the absolute change in the importance score of node $j$ when node $i$ is removed. Diagonal entries $M_{i,i}$ capture the self-dependence of each node, while large off-diagonal values reflect strong cross-node influence.

This metric offers several practical use cases. It enhances interpretability by identifying globally or locally dominant nodes within prediction explanations and detecting spurious correlations---such as when semantically irrelevant or constant nodes unduly influence outcomes. It supports model debugging by uncovering shortcut behaviors or fragile reasoning patterns and informs robustness analysis by guiding graph perturbations to assess model stability and sensitivity.

Next, we apply this proposed analysis across different benchmarks on the same vulnerable function used in Figure~\ref{fig:spurious_corr_code}, allowing visual comparison of attribution dynamics under varied training regimes.

The heatmap for the 100/0 benchmark reveals a narrow attribution dependency pattern, with notably high sensitivity centered around the variable \texttt{EPERM}. This suggests that the model disproportionately relies on this node when forming its prediction. However, this behavior likely reflects a spurious correlation, as \texttt{EPERM}---a standard error code---does not fundamentally determine the presence of a CWE-20 vulnerability. Conversely, critical vulnerability-relevant components such as \texttt{mode}, \texttt{profile\_mode\_names}, and \texttt{profile\_mode\_names[mode]}, which directly relate to the improper input validation flaw, show little to no influence on the attribution of other nodes. This indicates that the model fails to capture the semantic importance of these core vulnerability-inducing elements.

In contrast, the 50/50 benchmark, trained on a balanced mix of original and counterfactual examples, demonstrates a more structured and semantically aligned dependency map. Here, masking \texttt{profile\_mode\_names} or \texttt{profile\_mode\_names[mode]} significantly alters the attribution of related nodes, suggesting that the model has internalized the interdependence between these components. Furthermore, conditional statements and their associated return values also exhibit logical attribution interactions, implying better recognition of control-flow implications.

\emph{These observations underscore the effectiveness of the proposed node dependency metric in revealing both spurious and semantically meaningful attribution patterns.} The comparison between benchmarks demonstrates that counterfactual data augmentation improves the model’s learning consistency, encouraging reliance on true vulnerability signals rather than superficial correlations. Across all metrics, this collectively highlights counterfactual augmentation as a robust strategy for mitigating spurious learning and enhancing the stability, generalization, and interpretability of GNN-based vulnerability detection systems.

\section{Conclusions and Future Work}

The research presents a single unified framework, VISION, which detects vulnerabilities in source code through counterfactual data augmentation while providing interpretability. The method generates paired examples through systematic, minimal semantic changes to prevent models from learning spurious correlations that appear in noisy or imbalanced datasets. The model learns to detect actual vulnerability patterns rather than relying on superficial code features through GNN training on counterfactual pairs. The framework uses graph-based explainability to reveal important decision-making components while providing an interactive visualization module for human-in-the-loop analysis.

We also point out two limitations. First, VISION has been evaluated exclusively on CWE-20 in this study. However, extending VISION to new CWEs entails the same process: gather labeled examples, generate counterfactuals with an LLM, and translate the source code into CPGs. The downstream modules require no further adjustment. Second, the use of LLM-generated counterfactuals may occasionally introduce unrealistic or noisy modifications. Future work will therefore (i) evaluate the framework across a broader set of CWEs and programming languages to assess generalization; and (ii) integrate semantic-preserving generation approaches together with formal verification to assess the correctness of all generated counterfactuals. 

\section*{Acknowledgments}
D. Egea was a research intern at the University of Maryland, supported through a partnership between the Office of Global Engineering Leadership (OGEL) and Universidad Pontificia Comillas. B. Halder and S. Dutta were supported in part by Google Gift Funding and NSF CAREER Award No. 2340006. The authors also thank Yanjun Fu, Faisal Hamman, and Pasan Dissanayake for their valuable feedback and suggestions.

\bibliography{aaai25}

\end{document}